\documentclass{article}
\usepackage{spconf,amsmath,graphicx}
\usepackage{multirow}
\usepackage{amssymb}
\usepackage{algorithmic}
\usepackage[ruled,linesnumbered]{algorithm2e}

\title{ENHANCING GENERALIZATION OF INVISIBLE FACIAL PRIVACY CLOAK VIA GRADIENT ACCUMULATION}
\name{
	\begin{tabular}{c}
	  \it Xuannan Liu \qquad Yaoyao Zhong \qquad Weihong Deng\\ 
	  \it Hongzhi Shi \qquad Xingchen Cui \qquad Yunfeng Yin \qquad Dongchao Wen\thanks{*Corresponding author: Dongchao Wen, wendongchao@foxmail.com}$^*$
	\end{tabular}
}
\address{Inspur (Beijing) Electronic Information Industry Co., Ltd}

\usepackage{fancyhdr}
\begin{document}
	%
	\maketitle
	\begin{abstract}
		The blooming of social media and face recognition (FR) systems has increased people's concern about privacy and security. A new type of adversarial privacy cloak (class-universal) can be applied to all the images of regular users, to prevent malicious FR systems from acquiring their identity information. In this work, we discover the optimization dilemma in the existing methods -- the local optima problem in large-batch optimization and the gradient information elimination problem in small-batch optimization. To solve these problems, we propose Gradient Accumulation (GA) to aggregate multiple small-batch gradients into a one-step iterative gradient to enhance the gradient stability and reduce the usage of quantization operations. Experiments show that our proposed method achieves high performance on the Privacy-Commons dataset against black-box face recognition models.
	\end{abstract}

	\section{Introduction}
	\label{sec:intro}
	Recent advances in deep neural networks~\cite{he2016deep,howard2017mobilenets,hu2018squeeze} have improved the state of the art in face recognition (FR)~\cite{wang2018cosface,deng2018compressive,deng2019arcface,wang2020deep,zhong2021sface}. Unfortunately, with a huge amount of personal photos shared on social media, the abuse of FR systems has increased the potential risks of personal information leakage~\cite{yang2021towards,zhong2022opom,liu2023advcloak}. This makes face privacy protection critical.
	
\begin{figure}[!t]
		\centering
		\includegraphics[width=1.0\linewidth]{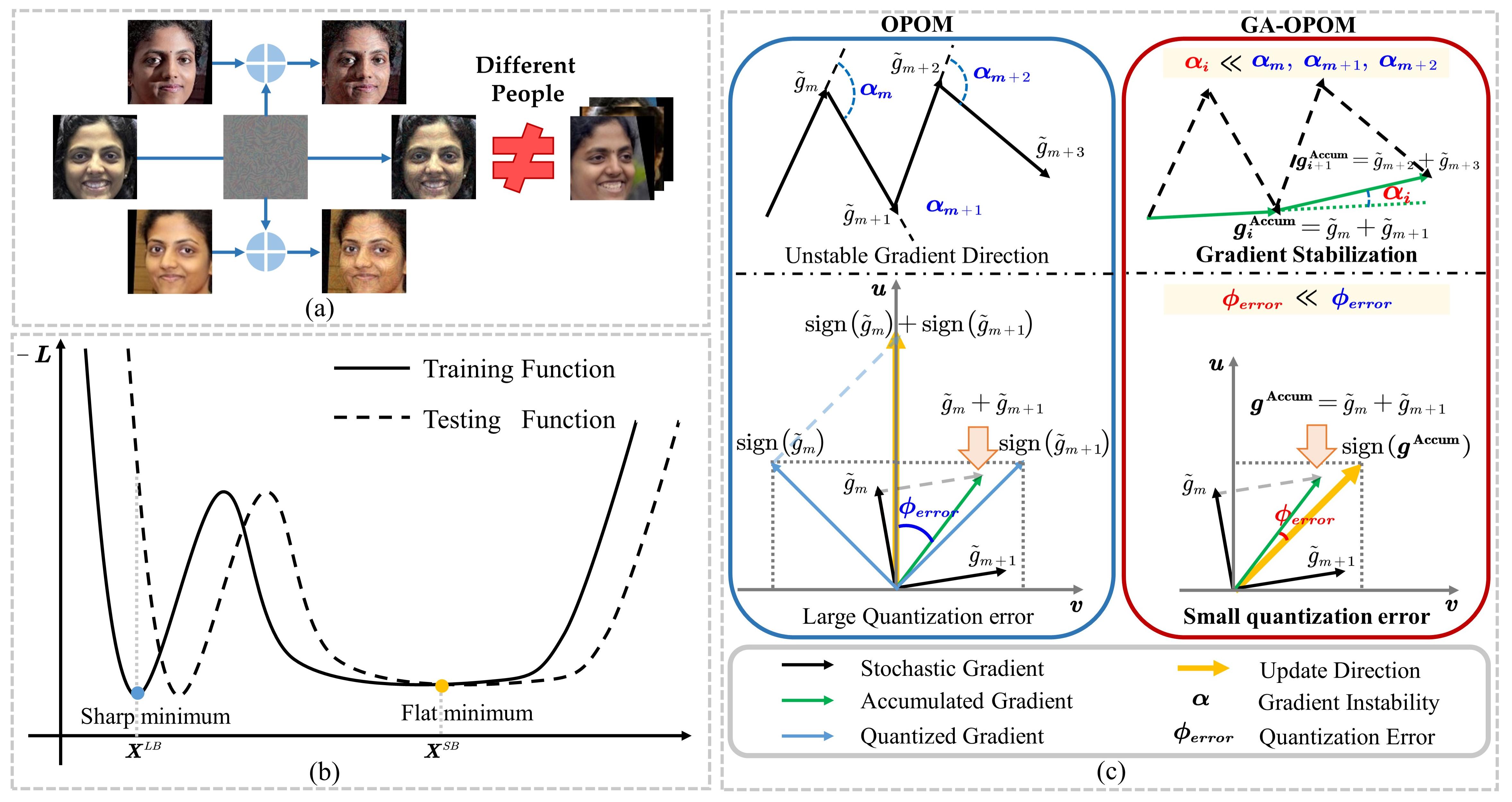}
		\caption{(a): Illustration of the person-specific (class-wise) privacy masks. (b): The flat and sharp minima in stochastic gradient optimization. (c): Two main issues, \textit{i.e.}, gradient instability and quantization error using small-batch training.}
		\label{gradient_vanish}
	\end{figure}

	Among the existing techniques, obfuscation techniques~\cite{sarwar2019privacy,fan2018image} have shown to be effective but at the cost of reducing the visual quality of the original images, thereby defeating the purpose of users sharing images on social media. Another group of works~\cite{shan2020fawkes,yang2021towards,cherepanova2021lowkey} employs adversarial techniques to hide people’s identities under tiny perturbations. However, the high computational cost makes it impractical to craft corresponding perturbations for seas of face images. Therefore, as shown in Fig. \ref{gradient_vanish} (a), a recent work~\cite{zhong2022opom} reveals the existence of person-specific (class-universal) adversarial masks that can protect all the face images of the same person and proposes an efficient generation method, OPOM. Due to the diversity of face images and FR models, the generalization of such privacy masks is more challenging.
	
	Similar with the training process of neural networks, the optimization of OPOM can be described as a non-convex problem and adopts the stochastic gradient algorithm. In the optimizing phase, the iterative gradients being acquired to traverse full training images can be viewed as the large-batch training. A previous work~\cite{keskar2016large} verifies that the large-batch training tends to converge to a sharp extremum leading to a large generalization gap in Fig. \ref{gradient_vanish} (b). Therefore, we encourage the privacy masks in optimization not only to achieve high attack loss but also to locate at flat regions. An intuitive idea is to exploit the inherent noise of gradient estimation to escape from the poor local optima by using small-batch training. However, we observe that OPOM optimized by small-batch methods has poor protection performance. Specifically, frequent use of sign operations in adversarial attacks will accumulate large amounts of quantization error~\cite{zhang2022revisiting}, especially combined with gradient instabilities, leading to the gradient information elimination problem, as shown in Fig. \ref{gradient_vanish} (c).
	
	In this work, we propose a gradient accumulation one person one mask method, termed GA-OPOM to further enhance the generalization ability of the privacy cloak. The key operation is to accumulate all small-batch gradients as a one-step iterative gradient for updating the adversarial perturbations. Such operation can effectively enhance the gradient stability and reduce the use of quantization operations while introducing the noisy gradients to escape from sharp local optima.
		
	\textbf{Contributions.} The contributions of our paper include:
	\begin{itemize}
		\setlength{\itemsep}{0pt} 
		\setlength{\parsep}{0pt}
		\setlength{\parskip}{0pt}
		\item We investigate the optimization dilemma behind the low generalization ability of the existing OPOM method -- the local optima problem using large-batch optimization, and the gradient information elimination problem using small-batch optimization.
		\item We propose the GA-OPOM method that stabilizes the gradient directions and reduces quantization errors to solve the optimization dilemma problems.
		\item We empirically demonstrate the effectiveness of our proposed method towards different black-box FR models compared with the SOTA methods.
	\end{itemize}
		
	\section{Preliminaries}
	\label{sec:Sys}
	Consider a variety of deep face recognition models $f\left( \cdot \right)$ and the face image set $X^k=\left\{ X_{1}^{k},...,X_{i}^{k},... \right\}$ of identity $k$. The objective of the OPOM privacy protection task is to find a perturbation $\Delta ^k$ for the identity $k$ satisfying:
	\begin{equation}
		\label{Opti}
		\begin{aligned}
			D\left( f\left( X_{i}^{k}+\Delta ^k \right) ,f_{X^k} \right) >t&,\quad ||\Delta ^k||_p\leqslant \epsilon,  \\
			for\,\,X_{i}^{k}\in X^k&,
		\end{aligned}
	\end{equation}
	where $f\left( X_{i}^{k} \right) \in \mathbb{R} ^d$ is the feature vector of $X_{i}^{k}$, $f_{X^k}$ denotes the feature subspace of identity $k$, $D\left( x_1,x_2 \right)$ measures the distance between $x_1$ and $x_2$, $t$ is the distance threshold to decide whether a pair of face images belong to the same identity, $\epsilon$ controls the magnitude of the adversarial mask. 
	
	To better describe the class-wise information of each identity, the affine hull is introduced to augment existing face images, by treating any affine combination of normalized deep features as a valid feature for the identity. Considering the feature subspace modeled by the affine hull may be excessively large, thus has a bad effect on the generation of person-specific masks. A direct thought is that we make upper and lower bound constraints on the weighting factor in the modeling of feature subspace. There exists a convex hull (the smallest convex set) when the lower bound is $0$, and the upper bound is $1$, which has been confirmed to be much tighter than the affine approximation~\cite{cevikalp2010face}, so that limited face images can realize their full potential,
	\begin{equation}
		\label{Con}
		H_{k}^{Con}=\left\{ x=\sum_{i=1}^n{\alpha _{i}^{k}f(X_{i}^{k})|\sum_{i=1}^n{\alpha _{i}^{k}=1},0\leqslant \alpha _{i}^{k}\leqslant 1} \right\}. 
	\end{equation}
	The convex hull $H_{k}^{Con}$ can be calculated through a least squares problem,
	\begin{equation}
		\label{least_square}
		\begin{aligned}
		\mathop {\mathrm{arg}\min}_{\alpha ^k}\left\| F^kA^k-f\left( X_{i}^{k}+\Delta ^k \right) \right\|, \\
		s.t.\,\,\sum{\alpha _{i}^{k}=1},\,\,0\leqslant \alpha _{i}^{k}\leqslant 1,
		\end{aligned}
	\end{equation}
	where $F^k$ is a matrix with columns $f(X_{i}^{k})$, and $A^k$ is a vector containing the corresponding coefficients ($\alpha _{i}^{k}$ of Equation \ref{Con}).
	
	\section{METHODOLOGY}
	\label{sec:metho}
	\subsection{Optimization Dilemma}
	Given that the optimization of OPOM can be formulated as a non-convex problem tackled through the stochastic gradient algorithm, it can be analogized to the training process of neural networks. In the OPOM method, the gradient estimation for each iteration needs to traverse the complete set of training images. However, a previous work~\cite{keskar2016large} confirms that the large-batch methods are easily attracted to regions with sharp extrema -- and as is well known, sharp extrema lead to poorer generalization. Similarly, when the adversarial privacy masks locate at sharp local extrema in large-batch optimization, the difference in face images and models will result in a significant change in attack loss, making the masks incapable of generalizing to unknown images and models. 
	
	On the other hand, the inherent noise within gradient estimation in the small-batch optimization can be exploited to escape from poor local optima. In small-batch methods, gradient instability~\cite{xiong2022stochastic,liu2023enhancing} is a normal concern due to the diversity in samples and models, and the sign function~\cite{goodfellow2014explaining,dong2018boosting,zhong2019adversarial,zhong2020towards} is a regular operation for efficiently generating adversarial examples. However, the sign operations could result in substantial optimization errors when iterative gradients exhibit severe instability. To illustrate this phenomenon, consider a simple toy example. Let $\tilde{g}_m$ and $\tilde{g}_{m+1}$ be gradients at iteration $m$ and $m+1$:
	\begin{equation}
		\tilde{g}_m=\left[ \cdots ,\underline{-0.01},0.10,0.25,\underline{1.00},\cdots \right] ^{\mathrm{T}} ,
	\end{equation}
	\begin{equation}
		\tilde{g}_{m+1}=\left[ \cdots ,\underline{1.00},0.05,0.10,\underline{-0.02},\cdots \right] ^T .
	\end{equation}
	When using the sign operations on the iterative gradients for updating privacy mask $\Delta$, a large value for the right region of $\tilde{g}_m$ is eliminated by the small negative value of $\tilde{g}_{m+1}$. Similarly in the left region, the sign operations lead to considerable deviations from the current optimization directions:
	\begin{equation}
		\mathrm{sign}\left( \tilde{g}_m \right) =\left[ \cdots ,\underline{-1},1,1,\underline{1},\cdots \right] ^T,
	\end{equation}
	\begin{equation}
		\mathrm{sign}\left( \tilde{g}_{m+1} \right) =\left[ \cdots ,\underline{1},1,1,\underline{-1},\cdots \right] ^T,
	\end{equation}
	\begin{equation}
		\begin{aligned}
			\Delta &=\mathrm{sign}\left( \tilde{g}_m \right) +\mathrm{sign}\left( \tilde{g}_{m+1} \right) \\
			&=\left[ \cdots ,\underline{0},2,2,\underline{0},\cdots \right] ^T .
		\end{aligned}
	\end{equation}
	The $\underline{0}$ represents the gradient information being eliminated, which is caused by the combination of gradient instability and sign operations.
	
	In summary, both the local optima problem using large-batch training and the gradient information elimination problem using small-batch training will hinder the generalization of adversarial privacy masks, resulting in the optimization dilemma of existing methods.
	
	\subsection{Attack algorithms}
	Therefore, our objective is to augment gradient stability and mitigate quantization errors in utilizing small-batch methods. 
	
	In this paper, we propose a simple strategy to accumulate multi-step noisy forward gradients into a one-step gradient update at each iteration in place of using the average gradient of full-batch face images in the OPOM method. Specifically, we first randomly select a face image $X_{m}^{k}$ repeatedly from the image set $\boldsymbol{X}^k$ to perform pre-search by updating the inner adversarial mask $\Delta ^{inner}$:
	\begin{equation}
		\tilde{g}_m=\nabla _{X_{adv}}\left( F^kA_{m}^{k}-f(X_{m}^{k}+\Delta _{m}^{inner}) \right), 
	\end{equation}
	\begin{equation}
		\Delta _{m+1}^{inner}=\mathrm{Clip}_{\epsilon}(\Delta _{m}^{inner}+sign(\tilde{g}_m)). 
	\end{equation}
	Then the inner stochastic gradients $\tilde{g}_m$ are accumulated for updating the outer iterative gradient $g^{\mathrm{Accum}}$:
	\begin{equation}
		g^{\mathrm{Accum}}\gets g^{\mathrm{Accum}}+\tilde{g}_m .
	\end{equation}
	After accumulating all the inner gradients into a one-step gradient of the outer iteration, we update the adversarial privacy mask $\Delta$ using the accumulated gradient:
	\begin{equation}
		\Delta _{N+1}=\mathrm{Clip}_{\epsilon}(\Delta _N+sign(g^{\mathrm{Accum}}) ,
	\end{equation}
	where $\mathrm{Clip}_{\epsilon}\left( \cdot \right)$ operation constrains the perturbation amplitude under the $l_{\infty}$ norm. 
	
	By accumulating critical gradient information, the iterative gradient estimates achieve high accuracy with a low variance simultaneously leading to a significant reduction in quantization operations. Moreover, the existence of noise gradients can help escape from poor local optima. The algorithm of GA-OPOM is summarized in Algorithm \ref{algorithm}.
	
	\begin{algorithm}[htbp]
		\caption{The GA-OPOM attack algorithm}
		\label{algorithm}
		\LinesNumbered
		\KwIn{Face images $\boldsymbol{X}^k = \left\{ X_1^k,X_2^k,\ldots ,X_{n_k}^k\right\}$ of identity $k$, deep face model $f(\cdot)$, maximum deviation of perturbations $\epsilon$, maximum iterative steps $N_{max}$.}	
		\textbf{Initialize:} $\Delta _0\sim U( -\epsilon ,\epsilon ) $, $N = 0$\;
		\While{step $N < N_{max}$}{
			$\Delta _{0}^{inner}=\Delta _N,\,\,g^{\mathrm{Accum}}=0$\;
			\For{$m=0\,\,to\,\,M-1$}
			{
				Random pick an image index $t\in \left\{ 1,...,n_k \right\}$\;
				Calculate $A_m^k$ by solving Equation~\ref{least_square}\;		
				
				$\tilde{g}_m=\nabla _{X_{adv}}\left( F^kA_{m}^{k}-f(X_{m}^{k}+\Delta _{m}^{inner}) \right) $\;
				
				$\Delta _{m+1}^{inner}=\mathrm{Clip}_{\epsilon}(\Delta _{m}^{inner}+sign(\tilde{g}_m))$\;
				$g^{\mathrm{Accum}}\gets g^{\mathrm{Accum}}+\tilde{g}_m$\;	
			}
			$\Delta _{N+1}=\mathrm{Clip}_{\epsilon}(\Delta _N+sign(g^{\mathrm{Accum}}))$\;	
			$N=N+1$\;}
		\KwOut{Privacy mask $\Delta _{N_{max}}$ for identity $k$.}	
	\end{algorithm}
	
	\begin{table*}[!h]
		\begin{center}
			\renewcommand{\arraystretch}{1.2}
			\resizebox{\linewidth}{!}{
				\begin{tabular}{c|c|cccccccccccc|cc}
					\hline
					\multirow{3}{*}{\textbf{\begin{tabular}[c]{@{}c@{}}Surrogate\\ Model\end{tabular}}} & \multirow{3}{*}{\textbf{Method}} & \multicolumn{12}{c|}{\textbf{Target Model}}                                                                                                                                                                                                                                                                                       & \multicolumn{2}{c}{\multirow{2}{*}{\textbf{\begin{tabular}[c]{@{}c@{}}Average Protection\\ Success Rate (\%)\end{tabular}}}} \\ \cline{3-14}
					&                                  & \multicolumn{2}{c|}{\textbf{ArcFace}}                & \multicolumn{2}{c|}{\textbf{CosFace}}                & \multicolumn{2}{c|}{\textbf{SFace}}                  & \multicolumn{2}{c|}{\textbf{MobileNet}}              & \multicolumn{2}{c|}{\textbf{SENet}}                  & \multicolumn{2}{c|}{\textbf{Inception-ResNet}} & \multicolumn{2}{c}{}                                                                                                         \\ \cline{15-16} 
					&                                  & \textbf{Top-1} & \multicolumn{1}{c|}{\textbf{Top-5}} & \textbf{Top-1} & \multicolumn{1}{c|}{\textbf{Top-5}} & \textbf{Top-1} & \multicolumn{1}{c|}{\textbf{Top-5}} & \textbf{Top-1} & \multicolumn{1}{c|}{\textbf{Top-5}} & \textbf{Top-1} & \multicolumn{1}{c|}{\textbf{Top-5}} & \textbf{Top-1}         & \textbf{Top-5}        & \textbf{Top-1}                                                & \textbf{Top-5}                                               \\ \hline \hline
					\multirow{4}{*}{ArcFace}                                                            & FI-UAP~\cite{zhong2022opom}                           & 82.3           & \multicolumn{1}{c|}{75.0}           & 71.2           & \multicolumn{1}{c|}{63.6}           & 77.3           & \multicolumn{1}{c|}{70.0}           & 65.2           & \multicolumn{1}{c|}{50.4}           & 73.9           & \multicolumn{1}{c|}{63.1}           & 56.6                   & 45.3                  & 71.1                                                          & 61.2                                                         \\
					& GA-FI-UAP\textbf{ (Ours)}                & \textbf{89.9}  & \multicolumn{1}{c|}{\textbf{85.4}}  & \textbf{82.2}  & \multicolumn{1}{c|}{\textbf{76.3}}  & \textbf{86.8}  & \multicolumn{1}{c|}{\textbf{81.2}}  & \textbf{77.8}  & \multicolumn{1}{c|}{\textbf{66.4}}  & \textbf{85.0}  & \multicolumn{1}{c|}{\textbf{77.8}}  & \textbf{67.3}          & \textbf{56.6}         & \textbf{81.5}                                                 & \textbf{73.9}                                                \\ \cline{2-16} 
					& OPOM-ConvexHull~\cite{zhong2022opom}                  & 86.5           & \multicolumn{1}{c|}{80.1}           & 76.8           & \multicolumn{1}{c|}{70.0}           & 82.7           & \multicolumn{1}{c|}{76.5}           & 70.5           & \multicolumn{1}{c|}{57.3}           & 79.3           & \multicolumn{1}{c|}{70.2}           & 63.2                   & 52.2                  & 76.5                                                          & 67.7                                                         \\
					& GA-OPOM-CenvexHull\textbf{ (Ours)}       & \textbf{91.1}  & \multicolumn{1}{c|}{\textbf{86.5}}  & \textbf{83.0}  & \multicolumn{1}{c|}{\textbf{77.3}}  & \textbf{87.5}  & \multicolumn{1}{c|}{\textbf{82.5}}  & \textbf{79.2}  & \multicolumn{1}{c|}{\textbf{68.2}}  & \textbf{85.8}  & \multicolumn{1}{c|}{\textbf{79.4}}  & \textbf{69.1}          & \textbf{58.9}         & \textbf{82.6}                                                 & \textbf{75.5}                                                \\ \hline \hline
					\multirow{4}{*}{CosFace}                                                            & FI-UAP~\cite{zhong2022opom}                           & 80.4           & \multicolumn{1}{c|}{72.3}           & 72.8           & \multicolumn{1}{c|}{65.1}           & 76.7           & \multicolumn{1}{c|}{68.9}           & 55.6           & \multicolumn{1}{c|}{41.0}           & 66.9           & \multicolumn{1}{c|}{54.6}           & 49.9                   & 37.1                  & 67.1                                                          & 56.5                                                         \\
					& GA-FI-UAP\textbf{ (Ours)}                & \textbf{90.0}  & \multicolumn{1}{c|}{\textbf{84.0}}  & \textbf{83.5}  & \multicolumn{1}{c|}{\textbf{77.4}}  & \textbf{87.0}  & \multicolumn{1}{c|}{\textbf{80.9}}  & \textbf{70.0}  & \multicolumn{1}{c|}{\textbf{56.9}}  & \textbf{80.4}  & \multicolumn{1}{c|}{\textbf{70.8}}  & \textbf{60.2}          & \textbf{47.9}         & \textbf{78.5}                                                 & \textbf{69.6}                                                \\ \cline{2-16} 
					& OPOM-ConvexHull~\cite{zhong2022opom}                  & 86.6           & \multicolumn{1}{c|}{79.3}           & 79.5           & \multicolumn{1}{c|}{72.7}           & 83.0           & \multicolumn{1}{c|}{76.3}           & 62.3           & \multicolumn{1}{c|}{48.2}           & 74.0           & \multicolumn{1}{c|}{63.1}           & 56.8                   & 44.5                  & 73.7                                                          & 64.0                                                         \\
					& GA-OPOM-CenvexHull\textbf{ (Ours)}       & \textbf{90.7}  & \multicolumn{1}{c|}{\textbf{85.6}}  & \textbf{84.8}  & \multicolumn{1}{c|}{\textbf{79.5}}  & \textbf{87.7}  & \multicolumn{1}{c|}{\textbf{82.9}}  & \textbf{71.9}  & \multicolumn{1}{c|}{\textbf{59.4}}  & \textbf{81.9}  & \multicolumn{1}{c|}{\textbf{73.5}}  & \textbf{63.3}          & \textbf{51.5}         & \textbf{80.1}                                                 & \textbf{72.1}                                                \\ \hline \hline
					\multirow{4}{*}{Softmax}                                                            & FI-UAP~\cite{zhong2022opom}                           & 72.3           & \multicolumn{1}{c|}{62.4}           & 63.5           & \multicolumn{1}{c|}{53.3}           & 70.3           & \multicolumn{1}{c|}{61.9}           & 73.9           & \multicolumn{1}{c|}{61.8}           & 77.4           & \multicolumn{1}{c|}{67.6}           & 52.4                   & 40.7                  & 68.3                                                          & 58.0                                                         \\
					& GA-FI-UAP\textbf{ (Ours)}                & \textbf{82.0}  & \multicolumn{1}{c|}{\textbf{74.3}}  & \textbf{75.1}  & \multicolumn{1}{c|}{\textbf{67.0}}  & \textbf{81.5}  & \multicolumn{1}{c|}{\textbf{74.0}}  & \textbf{84.4}  & \multicolumn{1}{c|}{\textbf{75.6}}  & \textbf{86.2}  & \multicolumn{1}{c|}{\textbf{79.7}}  & \textbf{61.4}          & \textbf{50.5}         & \textbf{78.4}                                                 & \textbf{70.2}                                                \\ \cline{2-16} 
					& OPOM-ConvexHull~\cite{zhong2022opom}                  & 78.0           & \multicolumn{1}{c|}{69.4}           & 70.2           & \multicolumn{1}{c|}{61.4}           & 76.1           & \multicolumn{1}{c|}{68.7}           & 79.2           & \multicolumn{1}{c|}{69.1}           & 82.9           & \multicolumn{1}{c|}{74.2}           & 58.7                   & 47.2                  & 74.2                                                          & 65.0                                                         \\
					& GA-OPOM-CenvexHull\textbf{ (Ours)}       & \textbf{82.3}  & \multicolumn{1}{c|}{\textbf{74.5}}  & \textbf{75.2}  & \multicolumn{1}{c|}{\textbf{68.0}}  & \textbf{81.4}  & \multicolumn{1}{c|}{\textbf{74.9}}  & \textbf{85.2}  & \multicolumn{1}{c|}{\textbf{76.3}}  & \textbf{86.7}  & \multicolumn{1}{c|}{\textbf{80.0}}  & \textbf{62.2}          & \textbf{51.4}         & \textbf{78.8}                                                 & \textbf{70.9}                                                \\ \hline
				\end{tabular}
			}
		\end{center}
		\caption{Comparison of different methods to generate person-specific privacy masks ($\epsilon =8$). We report the Top-1 and Top-5 protection success rate (\%) under the 1:N identification setting of the Privacy-Commons dataset.}
		\label{base_mf2}
	\end{table*} 
	
	\begin{table*}[!h]
		\begin{center}
			\renewcommand{\arraystretch}{1.2}
			\resizebox{\linewidth}{!}{
				\begin{tabular}{c|c|cccccccccccc|cc}
					\hline
					\multirow{3}{*}{\textbf{\begin{tabular}[c]{@{}c@{}}Surrogate\\ Model\end{tabular}}} & \multirow{3}{*}{\textbf{Method}} & \multicolumn{12}{c|}{\textbf{Target Model}}                                                                                                                                                                                                                                                                                       & \multicolumn{2}{c}{\multirow{2}{*}{\textbf{\begin{tabular}[c]{@{}c@{}}Average Protection\\ Success Rate (\%)\end{tabular}}}} \\ \cline{3-14}
					&                                  & \multicolumn{2}{c|}{\textbf{ArcFace}}                & \multicolumn{2}{c|}{\textbf{CosFace}}                & \multicolumn{2}{c|}{\textbf{SFace}}                  & \multicolumn{2}{c|}{\textbf{MobileNet}}              & \multicolumn{2}{c|}{\textbf{SENet}}                  & \multicolumn{2}{c|}{\textbf{Inception-ResNet}} & \multicolumn{2}{c}{}                                                                                                         \\ \cline{15-16} 
					&                                  & \textbf{Top-1} & \multicolumn{1}{c|}{\textbf{Top-5}} & \textbf{Top-1} & \multicolumn{1}{c|}{\textbf{Top-5}} & \textbf{Top-1} & \multicolumn{1}{c|}{\textbf{Top-5}} & \textbf{Top-1} & \multicolumn{1}{c|}{\textbf{Top-5}} & \textbf{Top-1} & \multicolumn{1}{c|}{\textbf{Top-5}} & \textbf{Top-1}         & \textbf{Top-5}        & \textbf{Top-1}                                                & \textbf{Top-5}                                               \\ \hline \hline
					\multirow{4}{*}{ArcFace}                                                            & FI-UAP~\cite{zhong2022opom}                           & 89.8           & \multicolumn{1}{c|}{84.9}           & 81.1           & \multicolumn{1}{c|}{75.3}           & 86.2           & \multicolumn{1}{c|}{81.0}           & 79.3           & \multicolumn{1}{c|}{68.5}           & 85.6           & \multicolumn{1}{c|}{78.8}           & 70.6                   & 60.9                  & 82.1                                                          & 74.9                                                         \\
					& GA-FI-UAP\textbf{ (Ours)}                & \textbf{95.2}  & \multicolumn{1}{c|}{\textbf{92.7}}  & \textbf{90.8}  & \multicolumn{1}{c|}{\textbf{87.1}}  & \textbf{93.4}  & \multicolumn{1}{c|}{\textbf{90.3}}  & \textbf{89.4}  & \multicolumn{1}{c|}{\textbf{83.0}}  & \textbf{92.9}  & \multicolumn{1}{c|}{\textbf{89.2}}  & \textbf{82.0}          & \textbf{74.7}         & \textbf{90.6}                                                 & \textbf{86.2}                                                \\ \cline{2-16} 
					& OPOM-ConvexHull~\cite{zhong2022opom}                  & 90.7           & \multicolumn{1}{c|}{86.2}           & 82.7           & \multicolumn{1}{c|}{77.0}           & 87.6           & \multicolumn{1}{c|}{82.7}           & 81.2           & \multicolumn{1}{c|}{71.3}           & 86.5           & \multicolumn{1}{c|}{80.6}           & 72.0                   & 63.0                  & 83.5                                                          & 76.8                                                         \\
					& GA-OPOM-CenvexHull\textbf{ (Ours)}       & \textbf{95.1}  & \multicolumn{1}{c|}{\textbf{92.8}}  & \textbf{90.9}  & \multicolumn{1}{c|}{\textbf{87.4}}  & \textbf{93.8}  & \multicolumn{1}{c|}{\textbf{91.2}}  & \textbf{90.2}  & \multicolumn{1}{c|}{\textbf{84.1}}  & \textbf{93.3}  & \multicolumn{1}{c|}{\textbf{89.5}}  & \textbf{82.0}          & \textbf{74.6}         & \textbf{90.9}                                                 & \textbf{86.6}                                                \\ \hline \hline
					\multirow{4}{*}{CosFace}                                                            & FI-UAP~\cite{zhong2022opom}                           & 88.6           & \multicolumn{1}{c|}{82.8}           & 82.5           & \multicolumn{1}{c|}{76.4}           & 85.5           & \multicolumn{1}{c|}{79.7}           & 69.7           & \multicolumn{1}{c|}{56.0}           & 79.2           & \multicolumn{1}{c|}{69.6}           & 62.0                   & 51.3                  & 77.9                                                          & 69.3                                                         \\
					& GA-FI-UAP\textbf{ (Ours)}                & \textbf{94.8}  & \multicolumn{1}{c|}{\textbf{92.1}}  & \textbf{91.4}  & \multicolumn{1}{c|}{\textbf{88.0}}  & \textbf{93.6}  & \multicolumn{1}{c|}{\textbf{90.2}}  & \textbf{84.7}  & \multicolumn{1}{c|}{\textbf{75.5}}  & \textbf{90.5}  & \multicolumn{1}{c|}{\textbf{85.3}}  & \textbf{76.5}          & \textbf{67.3}         & \textbf{88.6}                                                 & \textbf{83.0}                                                \\ \cline{2-16} 
					& OPOM-ConvexHull~\cite{zhong2022opom}                  & 89.6           & \multicolumn{1}{c|}{84.1}           & 84.2           & \multicolumn{1}{c|}{78.8}           & 87.8           & \multicolumn{1}{c|}{82.3}           & 73.0           & \multicolumn{1}{c|}{60.8}           & 82.0           & \multicolumn{1}{c|}{73.4}           & 64.8                   & 53.6                  & 80.2                                                          & 72.2                                                         \\
					& GA-OPOM-CenvexHull\textbf{ (Ours)}       & \textbf{95.0}  & \multicolumn{1}{c|}{\textbf{92.1}}  & \textbf{92.1}  & \multicolumn{1}{c|}{\textbf{88.6}}  & \textbf{94.0}  & \multicolumn{1}{c|}{\textbf{91.3}}  & \textbf{85.8}  & \multicolumn{1}{c|}{\textbf{77.0}}  & \textbf{90.9}  & \multicolumn{1}{c|}{\textbf{86.2}}  & \textbf{78.2}          & \textbf{68.5}         & \textbf{89.3}                                                 & \textbf{83.9}                                                \\ \hline \hline
					\multirow{4}{*}{Softmax}                                                            & FI-UAP~\cite{zhong2022opom}                           & 80.2           & \multicolumn{1}{c|}{71.2}           & 72.0           & \multicolumn{1}{c|}{63.9}           & 77.9           & \multicolumn{1}{c|}{70.9}           & 82.7           & \multicolumn{1}{c|}{72.9}           & 84.5           & \multicolumn{1}{c|}{77.2}           & 61.5                   & 50.9                  & 76.5                                                          & 67.8                                                         \\
					& GA-FI-UAP\textbf{ (Ours)}                & \textbf{88.4}  & \multicolumn{1}{c|}{\textbf{82.4}}  & \textbf{82.6}  & \multicolumn{1}{c|}{\textbf{76.6}}  & \textbf{87.6}  & \multicolumn{1}{c|}{\textbf{82.7}}  & \textbf{91.1}  & \multicolumn{1}{c|}{\textbf{85.0}}  & \textbf{91.5}  & \multicolumn{1}{c|}{\textbf{86.8}}  & \textbf{71.4}          & \textbf{61.9}         & \textbf{85.4}                                                 & \textbf{79.2}                                                \\ \cline{2-16} 
					& OPOM-ConvexHull~\cite{zhong2022opom}                  & 81.2           & \multicolumn{1}{c|}{72.6}           & 73.6           & \multicolumn{1}{c|}{65.7}           & 79.1           & \multicolumn{1}{c|}{72.2}           & 83.6           & \multicolumn{1}{c|}{74.6}           & 85.0           & \multicolumn{1}{c|}{78.0}           & 62.4                   & 52.1                  & 77.5                                                          & 69.2                                                         \\
					& GA-OPOM-CenvexHull\textbf{ (Ours)}       & \textbf{88.0}  & \multicolumn{1}{c|}{\textbf{82.0}}  & \textbf{82.9}  & \multicolumn{1}{c|}{\textbf{77.1}}  & \textbf{87.6}  & \multicolumn{1}{c|}{\textbf{82.7}}  & \textbf{91.0}  & \multicolumn{1}{c|}{\textbf{85.1}}  & \textbf{91.6}  & \multicolumn{1}{c|}{\textbf{86.7}}  & \textbf{71.4}          & \textbf{61.5}         & \textbf{85.4}                                                 & \textbf{79.2}                                                \\ \hline
				\end{tabular}
			}
		\end{center}
		\caption{Comparison of different methods combined with the momentum boosting method~\cite{dong2018boosting} and DFANet~\cite{zhong2020towards} to generate more transferable person-specific privacy masks ($\epsilon =8$). We report the Top-1 and Top-5 protection success rate (\%) under the 1:N identification setting of the Privacy-Commons dataset.}
		\label{base_mf2_DFANet}
	\end{table*} 
	
	\section{EXPERIMENTS}
	\label{sec:exper}
	\subsection{Image datasets, evaluation metrics and models}
	Following~\cite{zhong2022opom}, we use Privacy-Commons dataset, which consists of 500 individuals from the MegaFace challenge2 database~\cite{nech2017level}, each with 15 images (ten for training images and the other for testing images). For evaluation metrics, we report the Top-1 and Top-5 protection success rates (100\% - Top-1 or Top-5 accuracy). Three source models are the modified version~\cite{deng2019arcface} of ResNet-50, supervised by ArcFace~\cite{deng2019arcface}, CosFace~\cite{wang2018cosface} and Softmax loss. Besides, we choose six black-box models as the target models, three of which are different in the loss function, \textit{i.e.}, ArcFace, CosFace and SFace~\cite{zhong2021sface}, and the other are different in network structure, \textit{i.e.}, MobileNet~\cite{howard2017mobilenets}, SENet~\cite{hu2018squeeze} and Inception-ResNet (IncRes)~\cite{szegedy2017inception}.
	
	\subsection{Fair comparison and result analysis}
	Here we compared our method with two baseline methods, \textit{i.e.}, FI-UAP~\cite{zhong2022opom} and OPOM-ConvexHull~\cite{zhong2022opom}. FI-UAP only uses a single training point while OPOM-ConvexHull expands the source space to an appropriate degree. Some details are as follows. The maximum deviation $\epsilon$ is set to 8 and the number of iterative steps is chosen to be 16.
	
	As shown in Table \ref{base_mf2}, the gradient accumulation can significantly improve the protection success rate across all the models. For the privacy masks crafted on the CosFace model, the average protection success rate increases from 67.1\% and 73.7 \% to 78.5\% and 80.1\%, respectively. Such improvements verify GA-OPOM method can effectively enhance the cross-model generalization ability of the adversarial mask. 
	
	Model transferability methods can well improve protection performance across unknown models. We also evaluate the performance of the proposed GA-OPOM combined with transferability enhancement methods, \textit{i.e.}, momentum boosting~\cite{dong2018boosting} and DFANet~\cite{zhong2020towards}. Experimental results are listed in Table \ref{base_mf2_DFANet}. With the enhancement of the momentum method and DFANet, the privacy protection success rate of OPOM can increase further. In general, the GA-based methods consistently outperform the baseline methods by 7.4\% $\sim$ 10.7\% and exceed the current state-of-the-art method in attack performance, showing that the proposed method can be easily integrated with the existing model transferability methods.
	
	\begin{figure}[!ht]
		\centering
		\includegraphics[width=1.0\linewidth]{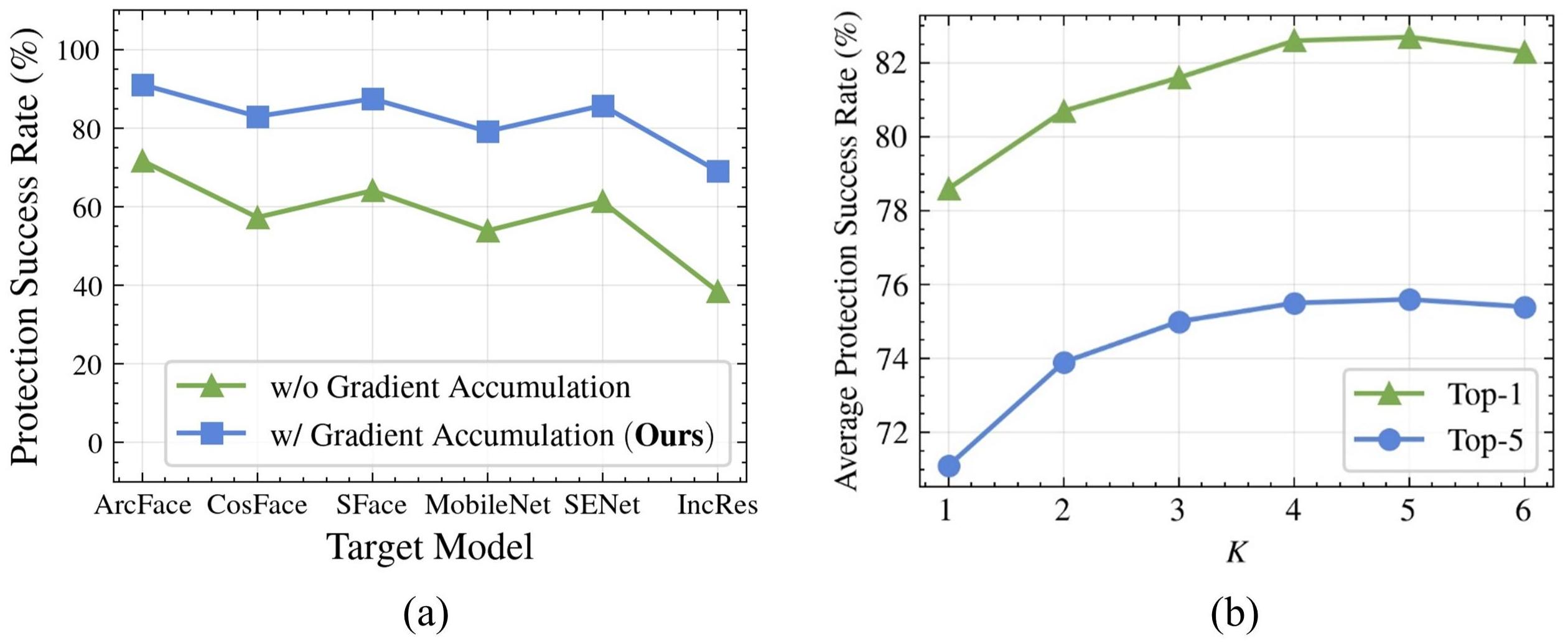}
		\caption{Ablation experiments to explore the impact of the gradient accumulation and the hyper-parameters for inner iteration number. (a): Ablation on the gradient accumulation. (b): Ablation on the inner iteration number.}
		\label{aba}
	\end{figure}

	\subsection{Ablation study}
	To show that the gains of gradient accumulation is not from increasing the number of gradient calculations, we report the results of OPOM-ConvexHull and GA-OPOM-ConvexHull with the equal total count of gradient calculations. The results in Fig. \ref{aba} (a) indicate that the improvement in protection performance comes from gradient accumulation.
	
	We continue to explore the effect of the inner iteration number $M$. We set the iteration number as $M=K\times \left| X^k \right|$. Fig. \ref{aba} (b) shows that as the the iteration number increases, the attack performance almost reaches the maximum when $K=4$. Choosing an appropriate iteration number can improve the generalization ability of the adversarial mask.
	
	\section{CONCLUSION}
	\label{sec:Conclu}
	In this paper, we explore the optimization dilemma in the class-wise universal adversarial perturbation, where the local optima problem using large-batch optimization, and the gradient information elimination problem using small-batch optimization. We tackle this problem by proposing gradient accumulation. GA-OPOM enables gradient queries to perform multiple rounds of inner pre-search by implementing small-batch optimization. Then all the gradients of the inner iterations are accumulated as a one-step gradient estimate and the accumulated gradients are used to update the outer adversarial perturbations. Experiments illustrate the superior performance of the proposed optimization methodology.
	
	\bibliographystyle{IEEEbib}
	\bibliography{strings,refs}

\begin{thebibliography}{10}

\bibitem{he2016deep}
Kaiming He, Xiangyu Zhang, Shaoqing Ren, and Jian Sun,
\newblock ``Deep residual learning for image recognition,''
\newblock in {\em CVPR}, 2016.

\bibitem{howard2017mobilenets}
Andrew~G Howard, Menglong Zhu, Bo~Chen, Dmitry Kalenichenko, Weijun Wang, Tobias Weyand, Marco Andreetto, and Hartwig Adam,
\newblock ``Mobilenets: Efficient convolutional neural networks for mobile vision applications,''
\newblock {\em arXiv preprint arXiv:1704.04861}, 2017.

\bibitem{hu2018squeeze}
Jie Hu, Li~Shen, and Gang Sun,
\newblock ``Squeeze-and-excitation networks,''
\newblock in {\em CVPR}, 2018.

\bibitem{wang2018cosface}
Hao Wang, Yitong Wang, Zheng Zhou, Xing Ji, Dihong Gong, Jingchao Zhou, Zhifeng Li, and Wei Liu,
\newblock ``Cosface: Large margin cosine loss for deep face recognition,''
\newblock in {\em CVPR}, 2018.

\bibitem{deng2018compressive}
Weihong Deng, Jiani Hu, and Jun Guo,
\newblock ``Compressive binary patterns: Designing a robust binary face descriptor with random-field eigenfilters,''
\newblock {\em PAMI}, 2018.

\bibitem{deng2019arcface}
Jiankang Deng, Jia Guo, Niannan Xue, and Stefanos Zafeiriou,
\newblock ``Arcface: Additive angular margin loss for deep face recognition,''
\newblock in {\em CVPR}, 2019.

\bibitem{wang2020deep}
Mei Wang and Weihong Deng,
\newblock ``Deep face recognition with clustering based domain adaptation,''
\newblock {\em Neurocomputing}, 2020.

\bibitem{zhong2021sface}
Yaoyao Zhong, Weihong Deng, Jiani Hu, Dongyue Zhao, Xian Li, and Dongchao Wen,
\newblock ``Sface: Sigmoid-constrained hypersphere loss for robust face recognition,''
\newblock {\em TIP}, 2021.

\bibitem{yang2021towards}
Xiao Yang, Yinpeng Dong, Tianyu Pang, Hang Su, Jun Zhu, Yuefeng Chen, and Hui Xue,
\newblock ``Towards face encryption by generating adversarial identity masks,''
\newblock in {\em ICCV}, 2021.

\bibitem{zhong2022opom}
Yaoyao Zhong and Weihong Deng,
\newblock ``Opom: Customized invisible cloak towards face privacy protection,''
\newblock {\em PAMI}, 2022.

\bibitem{liu2023advcloak}
Xuannan Liu, Yaoyao Zhong, Xing Cui, Yuhang Zhang, Peipei Li, and Weihong Deng,
\newblock ``Advcloak: Customized adversarial cloak for privacy protection,''
\newblock {\em arXiv preprint arXiv:2312.14407}, 2023.

\bibitem{sarwar2019privacy}
Omair Sarwar, Bernhard Rinner, and Andrea Cavallaro,
\newblock ``A privacy-preserving filter for oblique face images based on adaptive hopping gaussian mixtures,''
\newblock {\em IEEE Access}, 2019.

\bibitem{fan2018image}
Liyue Fan,
\newblock ``Image pixelization with differential privacy,''
\newblock in {\em IFIP Annual Conference on Data and Applications Security and Privacy}, 2018.

\bibitem{shan2020fawkes}
Shawn Shan, Emily Wenger, Jiayun Zhang, Huiying Li, Haitao Zheng, and Ben~Y Zhao,
\newblock ``Fawkes: Protecting privacy against unauthorized deep learning models,''
\newblock in {\em USENIX Security Symposium}, 2020.

\bibitem{cherepanova2021lowkey}
Valeriia Cherepanova, Micah Goldblum, Harrison Foley, Shiyuan Duan, John~P. Dickerson, Gavin Taylor, and Tom Goldstein,
\newblock ``Lowkey: Leveraging adversarial attacks to protect social media users from facial recognition,''
\newblock in {\em ICLR}, 2021.

\bibitem{keskar2016large}
Nitish~Shirish Keskar, Dheevatsa Mudigere, Jorge Nocedal, Mikhail Smelyanskiy, and Ping Tak~Peter Tang,
\newblock ``On large-batch training for deep learning: Generalization gap and sharp minima,''
\newblock in {\em ICLR}, 2017.

\bibitem{zhang2022revisiting}
Yihua Zhang, Guanhua Zhang, Prashant Khanduri, Mingyi Hong, Shiyu Chang, and Sijia Liu,
\newblock ``Revisiting and advancing fast adversarial training through the lens of bi-level optimization,''
\newblock in {\em ICML}, 2022.

\bibitem{cevikalp2010face}
Hakan Cevikalp and Bill Triggs,
\newblock ``Face recognition based on image sets,''
\newblock in {\em CVPR}, 2010.

\bibitem{xiong2022stochastic}
Yifeng Xiong, Jiadong Lin, Min Zhang, John~E Hopcroft, and Kun He,
\newblock ``Stochastic variance reduced ensemble adversarial attack for boosting the adversarial transferability,''
\newblock in {\em CVPR}, 2022.

\bibitem{liu2023enhancing}
Xuannan Liu, Yaoyao Zhong, Yuhang Zhang, Lixiong Qin, and Weihong Deng,
\newblock ``Enhancing generalization of universal adversarial perturbation through gradient aggregation,''
\newblock in {\em ICCV}, 2023.

\bibitem{goodfellow2014explaining}
Ian~J. Goodfellow, Jonathon Shlens, and Christian Szegedy,
\newblock ``Explaining and harnessing adversarial examples,''
\newblock in {\em ICLR}, 2015.

\bibitem{dong2018boosting}
Yinpeng Dong, Fangzhou Liao, Tianyu Pang, Hang Su, Jun Zhu, Xiaolin Hu, and Jianguo Li,
\newblock ``Boosting adversarial attacks with momentum,''
\newblock in {\em CVPR}, 2018.

\bibitem{zhong2019adversarial}
Yaoyao Zhong and Weihong Deng,
\newblock ``Adversarial learning with margin-based triplet embedding regularization,''
\newblock in {\em ICCV}, 2019.

\bibitem{zhong2020towards}
Yaoyao Zhong and Weihong Deng,
\newblock ``Towards transferable adversarial attack against deep face recognition,''
\newblock {\em TIFS}, 2020.

\bibitem{nech2017level}
Aaron Nech and Ira Kemelmacher-Shlizerman,
\newblock ``Level playing field for million scale face recognition,''
\newblock in {\em CVPR}, 2017.

\bibitem{szegedy2017inception}
Christian Szegedy, Sergey Ioffe, Vincent Vanhoucke, and Alexander~A Alemi,
\newblock ``Inception-v4, inception-resnet and the impact of residual connections on learning,''
\newblock in {\em AAAI}, 2017.

\end{thebibliography}
	
\end{document}